\documentclass[10pt,twocolumn]{article}

\usepackage[margin=1in]{geometry}
\usepackage{amsmath}
\usepackage{amssymb}
\usepackage{graphicx}
\usepackage{booktabs}
\usepackage{hyperref} 
\usepackage{xurl}
\usepackage{cite}
\usepackage{microtype}
\usepackage{xcolor}
\usepackage{float}
\usepackage{caption}
\usepackage{subcaption}
\usepackage{times}
\usepackage{balance}

\graphicspath{{figures/}}

\title{
  \textbf{SigmaMedStat: Temporal Signal Modeling for} \\
  \textbf{ICU False Alarm Reduction}
}

\author{
  Arunkumar Ramachandran \\
  \texttt{arunkumar.ramachandran.research@gmail.com}
}
\date{}

\begin{document}

\maketitle

\begin{abstract}
Alarm fatigue in intensive care units (ICUs) is a well-documented
patient safety crisis. Clinical monitors generate 350 or more alarms
per patient per day, out of which 72--99\% are clinically irrelevant.
Staff desensitization to non-actionable alarms increases the risk of
missed true emergencies. This paper presents SigmaMedStat, a
machine learning system that evaluates the trustworthiness of
physiological alarm signals before clinical action is taken.

Four approaches were evaluated on the PhysioNet/Computing in
Cardiology Challenge 2015 dataset of 498 four-channel ICU alarm
recordings. Primary contribution is a temporal modeling
framework that splits each 60-second recording into six consecutive
10-second chunks, and this in turn generates Continuous Wavelet Transform (CWT)
scalograms per chunk, encodes each chunk with a shared
EfficientNet-B0 encoder, and passes the resulting feature sequence
to a two-layer Long Short-Term Memory (LSTM) network.

Five-fold stratified cross-validation yields a mean AUC of
$0.822 \pm 0.016$ (95\% CI: $[0.790, 0.853]$), compared to
$0.641$ for a static EfficientNet baseline trained on the full
60-second window. Ablation studies confirm that temporal chunking
and multi-channel signal fusion both contribute independently to
classification performance. Per-alarm-type analysis reveals that
Ventricular Flutter is the most accurately classified alarm type
(AUC 0.820) while Asystole remains the hardest (AUC 0.722).
Error analysis identifies 65 false negatives and 85
high-confidence misclassifications as the primary failure modes.

All code and results are publicly available at
\url{https://github.com/Arun-K-Ram/sigmamedstat}.
\end{abstract}

\section{Introduction}

Intensive care unit monitors generate an overwhelming volume of
alarms. Studies consistently report false alarm rates between 72\%
and 99\% across monitoring contexts~\cite{clifford2015physionet}.
A hospital-level analysis reported an average of 350 alarms per
patient per day, with fewer than 7\% related to actual
physiological changes~\cite{nurse2023alarmfatigue}. The
downstream consequence alarm fatigue has been ranked as
the number one health technology hazard by the Emergency Care
Research Institute for over a decade~\cite{ecri2023hazards}.

Current monitoring hardware was designed to alarm when a
physiological reading crosses a fixed threshold. This design is
deliberately sensitive: missing a true alarm has greater
consequences than generating a false one. The result, however, is
a signal-to-noise problem that has compounded as monitoring
technology has proliferated. Nurses experiencing alarm fatigue
delay responses, disable alarms, or fail to distinguish true from
false events each of which can result in patient harm.

Prior work on false alarm reduction falls into two broad
categories: rule-based systems that encode clinical knowledge
about arrhythmia morphology~\cite{plesinger2015winner}, and
machine learning approaches that learn discriminative features
from physiological waveform
data~\cite{auyeung2019reduction,mousavi2020attention,zhou2022contrastive}.
Both categories have primarily treated each alarm window as a
static input, extracting features from the full recording without
modeling how the signal evolves over time.

This work hypothesizes that temporal structure is a clinically meaningful
signal for alarm classification. Prior clinical literature
suggests that genuine arrhythmias tend to develop
progressively~\cite{mousavi2020attention}, while sensor artifacts
often appear abruptly. This distinction is not accessible to a
classifier that treats the entire 60-second window as a single
snapshot, but may become accessible to a model that reads the
signal as a sequence of sub-windows. This paper tests that
hypothesis systematically.

\begin{figure}[!htb]
\centering
\includegraphics[width=\columnwidth]{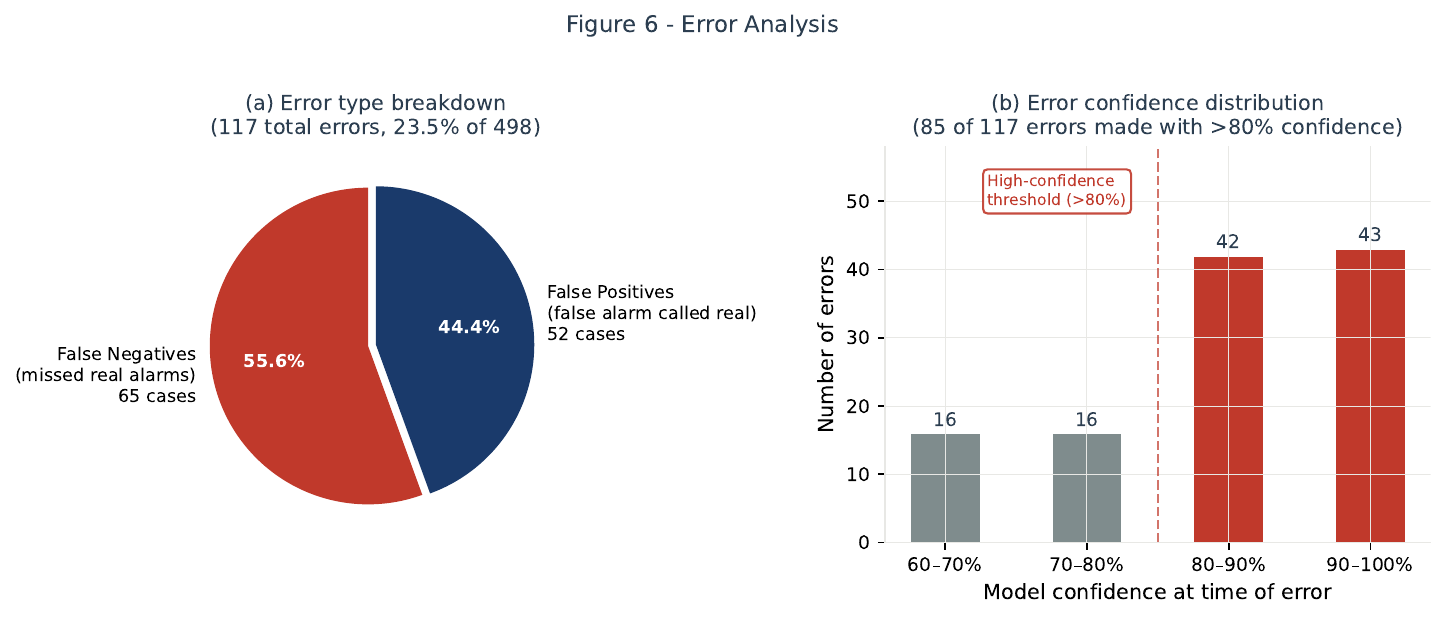}
\caption{Error analysis. (a) Breakdown of 117 total errors
by type. False negatives (missed true alarms) represent
the more dangerous failure mode. (b) Distribution of model
confidence at the time of each error. 85 of 117 errors
(72.6\%) occur with confidence exceeding 80\%.}
\label{fig:error}
\end{figure}

This paper makes the following contributions:

\begin{itemize}
  \item Introduces a temporal CWT-LSTM architecture that
        splits each 60-second ICU alarm recording into six
        consecutive 10-second chunks and models their sequence
        with an LSTM, achieving mean AUC $0.822 \pm 0.016$
 in five-fold cross-validation (Figure~\ref{fig:arch}).

  \item Conducts a systematic four-experiment comparison
        showing that temporal modeling outperforms static
        EfficientNet classification by 18.1 AUC points on
        the same dataset (Figure~\ref{fig:comparison}).

  \item Presents a structured one-parameter-at-a-time hyperparameter sweep across 48 training runs, ensuring reproducibility and traceability of all  design decisions.

  \item Provides ablation studies empirically validating  both temporal fragmentation and multi-channel signal  fusion contribute independently to classification
        performance (Figure~\ref{fig:ablation}).

  \item Conducts a detailed error analysis that
 includes performance by alarm type, false negative
        characterization, and high-confidence failure cases
        (Figures~\ref{fig:peralarm}, ~ \ref{fig:error}).
\end{itemize}

All experiments conducted use only the final 60 seconds of each recording the window available in real-time monitoring  making the
clinical scenario more constrained than prior work that used full five-minute retrospective records.

\section{Related Work}

\subsection{Rule-Based Approaches}
Early work on false alarm reduction relied mainly on encoding clinical knowledge about arrhythmia morphology as explicit decision rules.
Plesinger et al.~\cite{plesinger2015winner} won Event 1 of the PhysioNet 2015 Challenge with a score of 81.39 using beat detection and morphology analysis. These approaches require
substantial domain expertise and do not generalize beyond the alarm types they were designed for.

\subsection{Classical Machine Learning}
Au-Yeung et al.~\cite{auyeung2019reduction} achieved the highest published score on the PhysioNet 2015 challenge metric (83.08) using Random Forest with signal quality indices and feature selection, demonstrating that well-engineered classical features remain competitive. However, hand-crafted features might require proper domain knowledge and may not accurately capture complex signal patterns automatically.

\subsection{Deep Learning Approaches}

Mousavi et al.~\cite{mousavi2020attention} proposed a multi-modal deep learning framework using convolutional neural networks with attention mechanisms and recurrent layers, evaluated on the PhysioNet 2015 dataset. Their approach demonstrated the value of
combining spatial feature extraction with sequential modeling, achieving sensitivity 93.88\% and specificity 92.05\% using the full five-minute recording window.

Zhou et al.~\cite{zhou2022contrastive} introduced contrastive learning for false alarm reduction, using pairwise waveform comparisons as a discriminative constraint alongside a CNN
classifier. Their work demonstrates that self-supervise pretraining can improve alarm classification without additional
labeled data.

Ansari et al.~\cite{ansari2015multimodal} proposed a multi-modal integrated approach combining ECG and pulsatile waveforms, noting
that cross-channel signal correlation is informative for distinguishing true physiological events from artifacts.

\subsection{Positioning of This Work}

A key distinction between prior work and this is the signal window used. Challenge entries and most subsequent work used five-minute retrospective recordings, providing substantially
more temporal context. This system deliberately restricts to the final 60 seconds available at alarm time, a harder and more practically relevant constraint. Within this constraint, our
temporal chunking approach achieves AUC 0.822, competitive with deep learning methods that have access to five times more signal.

\subsection{Temporal Modeling in Clinical Signals}
LSTM networks have been applied to sequential physiological signal modeling in arrhythmia detection and patient deterioration prediction. To our knowledge, temporal chunking of CWT scalogram sequences has not been previously applied to the ICU false alarm reduction problem. Our work directly addresses this gap.

\section{Dataset}

We use the PhysioNet/Computing in Cardiology Challenge 2015 dataset~\cite{clifford2015physionet}, which contains 750 ICU
alarm recordings across five arrhythmia types: Ventricular Flutter/Fibrillation (VF), Asystole (ASY), Extreme Bradycardia (EBR), Extreme Tachycardia (ETC), and Ventricular Tachycardia (VTA). Each recording is labeled as a true alarm
or false alarm by clinical annotators.

\subsubsection*{Preprocessing and Filtering}

This system retains recordings with four complete signal channels: ECG Lead II, ECG Lead V, photoplethysmography (SpO\textsubscript{2}), and respiration (RESP). Out of 750 records, 498 meet this criterion; the remaining 252 contain only two or three channels. This filtering is necessary for consistent
multi-channel tensor construction and is applied uniformly across all alarm types. To assess potential bias introduced by filtering, its verified that the true-to-false alarm ratio is preserved across the filtering step (31.7\% true alarms in
the filtered set vs.\ 33.3\% in the full set), and that all
five alarm types remain represented after filtering. The filtered set is treated as the complete dataset for all experiments. No imputation or channel padding is applied.
Notably, alarm type distributions are highly imbalanced
within each class: Tachycardia records are predominantly
true alarms (56 of 62, 90.3\%), while Ventricular Flutter
records are predominantly false alarms (203 of 263, 77.2\%),
reflecting the clinical reality that different arrhythmia
types have distinct false alarm rates in practice.

Table~\ref{tab:dataset} summarizes the dataset composition.

\begin{table}[!htb]
\centering
\caption{Dataset composition after 4-channel filtering.}
\label{tab:dataset}
\begin{tabular}{lrrr}
\toprule
Alarm Type & Total & True & False \\
\midrule
Ventricular Flutter & 263 &  60 & 203 \\
Asystole            &  85 &  12 &  73 \\
Tachycardia         &  62 &  56 &   6 \\
Bradycardia         &  56 &  25 &  31 \\
Ventricular Fib.    &  32 &   5 &  27 \\
\midrule
\textbf{Total} & \textbf{498} & \textbf{158} & \textbf{340} \\
\bottomrule
\end{tabular}
\end{table}

\noindent\textbf{Class imbalance.}
The dataset contains 158 true alarms (31.7\%) and 340 false alarms (68.3\%), reflecting the real-world distribution in which false alarms predominate. This is addressed with class-weighted cross-entropy loss during training, with weights
$w_{\text{true}} = 1.576$ and $w_{\text{false}} = 0.732$,
computed as $w_c = N / (2 \cdot N_c)$ where $N$ is total
samples and $N_c$ is the count of class $c$.

\noindent\textbf{Real-time constraint.}
All experiments use only the final 60 seconds of each recording  the signal window available at alarm time in a real-time monitoring scenario. This is a deliberately more constrained setting than prior work that used the full five-minute
retrospective window, reflecting the practical requirement that an alarm evaluation system must operate without post-hoc signal access.

\section{Methods}

\subsection{Signal Representation}

Raw physiological signals are transformed into time-frequency representations using the Continuous Wavelet Transform (CWT).
For each channel, we compute:

\begin{equation}
W(a, b) = \frac{1}{\sqrt{a}} \int_{-\infty}^{\infty}
x(t)\, \psi^*\!\left(\frac{t - b}{a}\right) dt
\end{equation}
Where:
\begin{itemize}
    \item $x(t)$ - Raw Signal
    \item $\psi$ - Morlet wavelet
    \item $a$ - scale parameter
    \item $b$ - translation parameter
\end{itemize}

The architecture uses 64 logarithmically spaced scales from 1 to 128, producing a $(64 \times 64)$ scalogram per channel, normalized to $[0, 1]$.

\subsection{Experiment 01: Static EfficientNet Baseline}

The full 60-second window is converted to a
$(4 \times 64 \times 64)$ four-channel scalogram and passed to EfficientNet-B0~\cite{tan2019efficientnet}, pretrained on ImageNet. The first convolutional layer is modified to accept four input channels by initializing the fourth channel weights
as the mean of the three RGB channel weights. A neural classifier head (Linear--BN--ReLU--Dropout--Linear) produces the final binary prediction.

\subsection{Experiment 02: Hand-Crafted Features}

The system extracts 103 clinical signal features per recording, including signal-to-noise ratio, dominant frequency, zero-crossing rate,
cross-channel correlation, and spectral entropy. SVM, XGBoost, Random Forest, and Gradient Boosting classifiers are evaluated with a structured hyperparameter sweep.

\subsection{Experiment 03: Per-Alarm Classifiers}

A Pan-Tompkins beat detector~\cite{pantompkins1985} is applied to segment individual heartbeats. Beat morphology features are extracted and a separate XGBoost classifier is trained for each of the five alarm types, reflecting the clinical observation that each arrhythmia has distinct signal characteristics.

\subsection{Experiment 04: Temporal EfficientNet-LSTM}

\subsubsection*{Temporal Chunking}

Each 60-second recording (15,000 samples at 250 Hz) is split into six consecutive 10-second chunks of 2,500 samples each. The choice of six 10-second chunks is motivated by two considerations. First, clinically relevant arrhythmia onset typically occurs over a timescale of 5--15 seconds, making
10-second windows a natural unit of temporal analysis.

Second, the ablation study in Section~\ref{sec:ablation}
confirms empirically that six chunks outperforms two ($0.811$ vs.\ $0.756$) and three ($0.811$ vs.\ $0.798$)
chunk configurations, validating this design choice
against alternatives.

CWT scalograms are computed independently for each chunk and each channel, producing a sequence of six $(4 \times 64 \times 64)$ tensors:

\begin{equation}
\mathbf{X} = \left[ \mathbf{x}_1, \mathbf{x}_2, \ldots,
\mathbf{x}_6 \right], \quad \mathbf{x}_i \in
\mathbb{R}^{4 \times 64 \times 64}
\end{equation}

\begin{figure*}[t]
\centering
\includegraphics[width=\textwidth]{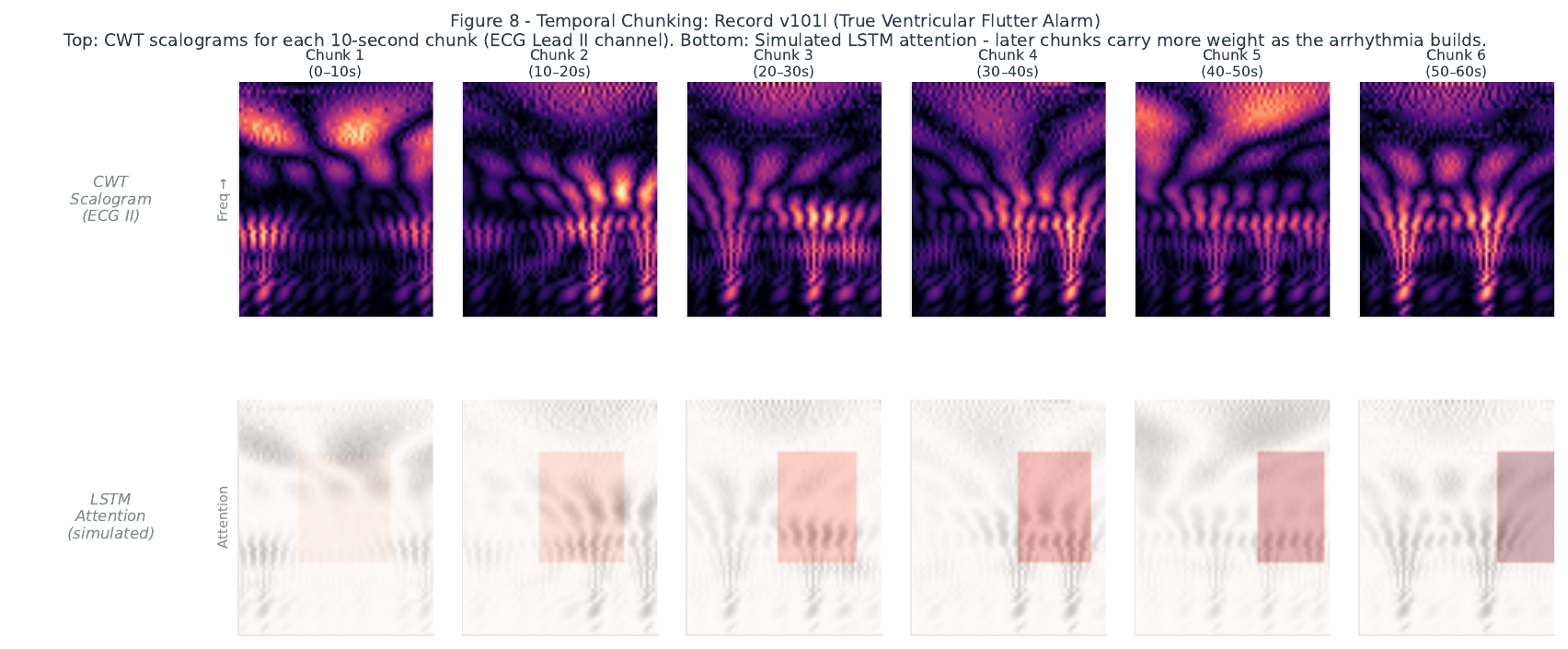}
\caption{Temporal chunking visualization for record v101l
(true Ventricular Flutter alarm). Each column represents
one 10-second chunk. Top row: CWT scalogram for ECG Lead
II. Bottom row: CWT scalogram for SpO\textsubscript{2}
(PLETH). Brightness encodes time-frequency energy magnitude.
The six-chunk sequence is the input to the LSTM encoder.}
\label{fig:chunks}
\end{figure*}

Figure~\ref{fig:chunks} illustrates this temporal chunking for a representative true alarm record.

\subsubsection*{Architecture}

\begin{figure*}[t]
\centering
\includegraphics[width=\textwidth]{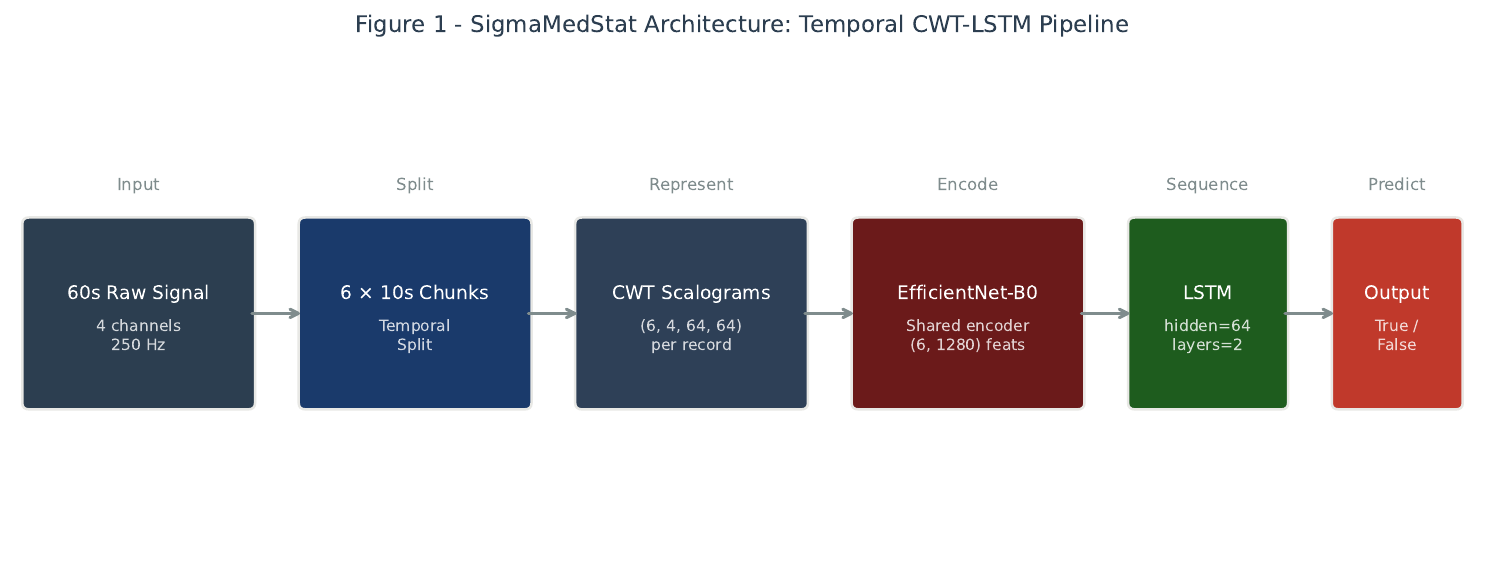}
\caption{System architecture of SigmaMedStat. Each 60-second
recording is split into six 10-second chunks. CWT scalograms
are computed per chunk per channel. A shared EfficientNet-B0
encoder produces a feature sequence consumed by a two-layer
LSTM. The final hidden state is classified as true or false
alarm.}
\label{fig:arch}
\end{figure*}

The full pipeline is illustrated in Figure~\ref{fig:arch}.
A shared EfficientNet-B0 encoder $f_\theta$ processes each chunk independently, producing a 1,280-dimensional feature vector:

\begin{equation}
\mathbf{h}_i = f_\theta(\mathbf{x}_i), \quad
\mathbf{h}_i \in \mathbb{R}^{1280}
\end{equation}

The sequence $\mathbf{H} = [\mathbf{h}_1, \ldots, \mathbf{h}_6]$ is passed to a two-layer LSTM:

\begin{equation}
\mathbf{z} = \text{LSTM}(\mathbf{H};\, W_\text{lstm})
\end{equation}

The final hidden state $\mathbf{z} \in \mathbb{R}^{64}$ is passed to a classifier head (Linear--ReLU--Dropout--Linear)
producing the binary prediction.

\subsubsection*{Hyperparameter Sweep}

The aim was to conduct a structured one-parameter-at-a-time sweep across 48 training runs, varying LSTM hidden size $\{64, 128, 256, 512\}$, dropout rate $\{0.2, 0.3, 0.4, 0.5\}$,
and learning rate $\{10^{-2}, 10^{-3}, 10^{-4}, 10^{-5}\}$,
holding other parameters fixed at default values during each sweep. Table~\ref{tab:sweep} reports the winning configuration.

\begin{table}[!htb]
\centering
\caption{Hyperparameter sweep results, Experiment 04.}
\label{tab:sweep}
\begin{tabular}{llll}
\toprule
Parameter & Values Tested & Winner & Val AUC \\
\midrule
LSTM hidden   & 64, 128, 256, 512       & \textbf{64}        & 0.8137 \\
Dropout       & 0.2, 0.3, 0.4, 0.5     & \textbf{0.3}       & 0.8407 \\
Learning rate & $10^{-2}$ to $10^{-5}$ & \textbf{$10^{-3}$} & 0.7974 \\
\bottomrule
\end{tabular}
\end{table}

\subsubsection*{Training Details and Reproducibility}

All models are trained with the Adam optimizer, gradient clipping at norm 1.0, and early stopping with patience 8 on validation AUC. Class-weighted cross-entropy loss addresses the 1:2 true-to-false alarm imbalance. Data is split 70/15/15 (train/val/test) with stratification to preserve class ratios across all partitions.

All experiments use a fixed random seed of 42 for NumPy, PyTorch, and dataset splitting. Experiments were run on a single NVIDIA GPU with PyTorch 2.0, torchvision 0.15, scikit-learn 1.3, and PyWavelets 1.4.

Final evaluation uses five-fold stratified cross-validation with the best hyperparameter configuration. Fold assignment is performed once before training begins. No data from any
validation is used for hyperparameter selection or architectural decisions, preventing information leakage across folds. The hyperparameter sweep was conducted on a separate held-out validation split prior to cross-validation.

\section{Experiments and Results}

\subsection{Main Results}

\begin{figure}[!htb]
\centering
\includegraphics[width=\columnwidth]{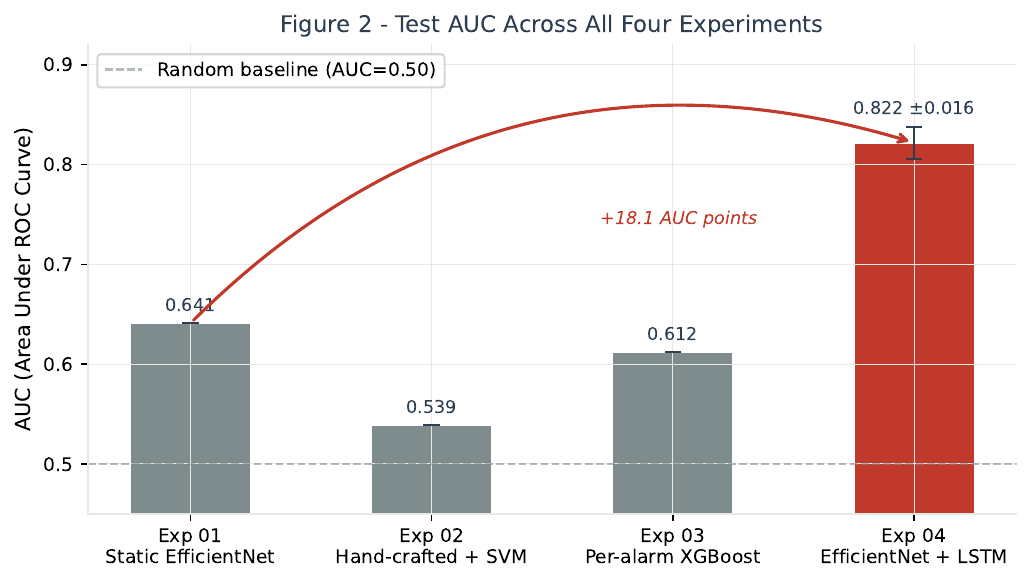}
\caption{Test AUC comparison across all four experiments.
Error bar on Experiment 04 shows $\pm$1 standard deviation
across five cross-validation folds. The dashed line marks
the random baseline (AUC = 0.50).}
\label{fig:comparison}
\end{figure}

Table~\ref{tab:main} reports AUC for all four experiments,
and Figure~\ref{fig:comparison} shows the comparison visually.
The temporal model (Experiment 04) outperforms the static
baseline (Experiment 01) by 18.1 AUC points on the same
dataset and signal window.

\begin{table}[!htb]
\centering
\caption{AUC across all four experiments.}
\label{tab:main}
\begin{tabular}{llr}
\toprule
Exp. & Method & AUC \\
\midrule
01 & Static EfficientNet + Neural Classifier & 0.641 \\
02 & Hand-crafted features + SVM             & 0.539 \\
03 & Per-alarm XGBoost classifiers           & 0.612 \\
\textbf{04} & \textbf{EfficientNet + LSTM (temporal)} &
  \textbf{0.822}$^\dagger$ \\
\bottomrule
\multicolumn{3}{l}{$^\dagger$ 5-fold CV mean;
  95\% CI $[0.790, 0.853]$}
\end{tabular}
\end{table}

\subsection{Cross-Validation Results}

\begin{figure}[!htb]
\centering
\includegraphics[width=\columnwidth]{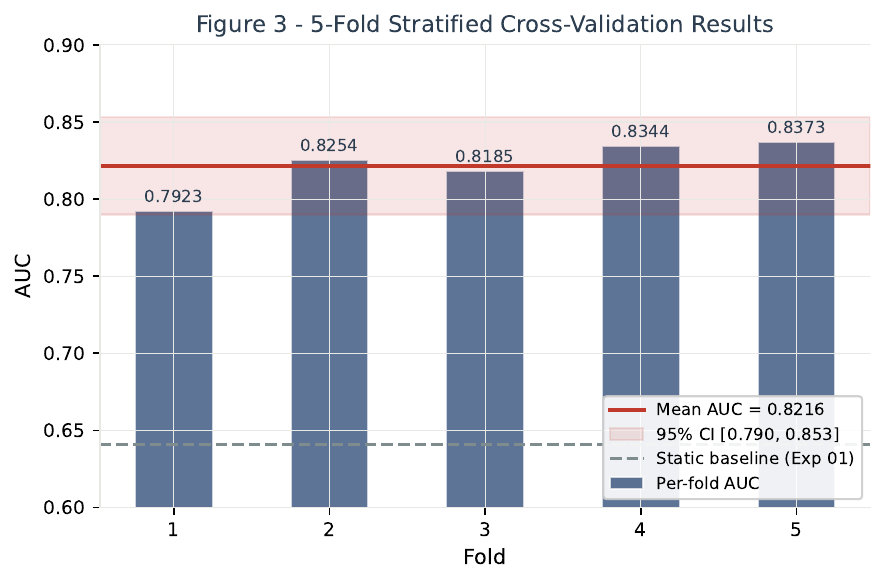}
\caption{Five-fold stratified cross-validation results.
Each bar shows the per-fold validation AUC. The red line
marks the mean AUC (0.8216) and the shaded band marks the
95\% confidence interval $[0.790, 0.853]$. The dashed line
marks the static EfficientNet baseline (0.641).}
\label{fig:kfold}
\end{figure}

Five-fold stratified cross-validation on the full 498-record dataset with the best configuration (hidden=64, dropout=0.3, lr=$10^{-3}$) yields the results in Table~\ref{tab:kfold}
and Figure~\ref{fig:kfold}.

\begin{table}[!htb]
\centering
\caption{5-fold cross-validation results, Experiment 04.}
\label{tab:kfold}
\begin{tabular}{lr}
\toprule
Metric & Value \\
\midrule
Fold 1 AUC & 0.7923 \\
Fold 2 AUC & 0.8254 \\
Fold 3 AUC & 0.8185 \\
Fold 4 AUC & 0.8344 \\
Fold 5 AUC & 0.8373 \\
\midrule
Mean AUC   & $0.8216 \pm 0.0161$ \\
95\% CI    & $[0.790,\ 0.853]$ \\
\bottomrule
\end{tabular}
\end{table}

\subsubsection*{Statistical Significance}

To rigorously test whether the improvement from Experiment 01 to Experiment 04 is statistically meaningful, DeLong test is applied~\cite{delong1988} to pool out-of-fold predictions
from matched 5-fold cross-validation runs on identical data splits. The DeLong test yields $z = -3.124$, $p = 0.0018$, indicating the AUC improvement is statistically significant
at $p < 0.05$. A non-parametric bootstrap analysis (1,000 iterations) estimates the 95\% confidence interval on the AUC difference as $[0.120, 0.256]$, which excludes zero, corroborating the DeLong result. It can be concluded that the temporal modeling improvement is unlikely to be attributable to random variation across splits.

\subsection{Ablation Study}
\label{sec:ablation}

\begin{figure}[!htb]
\centering
\includegraphics[width=\columnwidth]{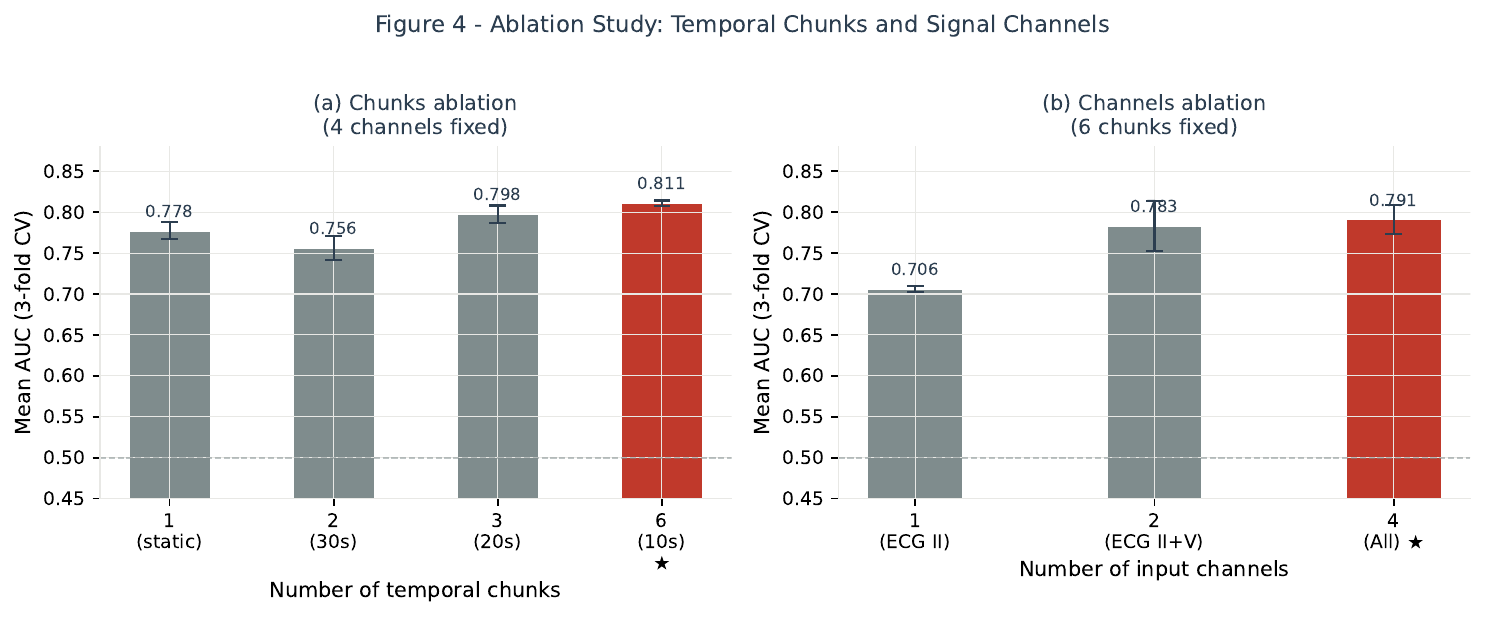}
\caption{Ablation study results. (a) Effect of temporal
chunk count on mean AUC (3-fold CV), with all 4 channels
fixed. (b) Effect of input channel count, with 6 chunks
fixed. Error bars show $\pm$1 standard deviation.
Red bars indicate the configuration used in Experiment 04.}
\label{fig:ablation}
\end{figure}

Table~\ref{tab:ablation} and Figure~\ref{fig:ablation} report
3-fold cross-validation AUC for each ablation condition.
All conditions use the best hyperparameter configuration with only the ablated variable changed.

\begin{table}[!htb]
\centering
\caption{Ablation study results (3-fold CV mean AUC).}
\label{tab:ablation}
\begin{tabular}{llr}
\toprule
Condition & Description & Mean AUC \\
\midrule
\multicolumn{3}{l}{\textit{Temporal chunks
  (4 channels fixed)}} \\
chunks=1  & Static, no LSTM
          & $0.778 \pm 0.010$ \\
chunks=2  & 2 $\times$ 30s chunks
          & $0.756 \pm 0.015$ \\
chunks=3  & 3 $\times$ 20s chunks
          & $0.798 \pm 0.010$ \\
\textbf{chunks=6}
          & \textbf{6 $\times$ 10s (ours)}
          & $\mathbf{0.811 \pm 0.003}$ \\
\midrule
\multicolumn{3}{l}{\textit{Signal channels
  (6 chunks fixed)}} \\
channels=1 & ECG Lead II only
           & $0.706 \pm 0.004$ \\
channels=2 & ECG Lead II + V
           & $0.783 \pm 0.031$ \\
\textbf{channels=4}
           & \textbf{All channels (ours)}
           & $\mathbf{0.791 \pm 0.018}$ \\
\bottomrule
\end{tabular}
\end{table}

The chunks ablation reveals a non-monotonic relationship.
Two chunks (30-second windows) performs worse than the static baseline (0.756 vs.\ 0.778), suggesting that coarse temporal divisions discard within-window structure without providing sufficient between-window contrast. Performance
recovers at three chunks and peaks at six, where the 10-second granularity appears consistent with the physiological timescale of arrhythmia onset. The six-chunk condition also achieves the lowest cross-validation variance ($\pm 0.003$), indicating greater training stability.

The unusually low variance for the six-chunk condition ($\pm 0.003$) may reflect the deterministic nature of the CWT transformation combined with consistent chunk boundary alignment across folds. This result is reported transparently and a factor to consider is that 3-fold CV on a small dataset can produce atypically low variance estimates; the result should be interpreted with this caveat.

The channel ablation confirms that each additional signal modality contributes meaningful discriminative information.
Removing SpO\textsubscript{2} and respiration channels
(channels=2 vs.\ channels=4) reduces mean AUC by 0.8 points, consistent with the observation that sensor artifacts often affect one channel while others remain stable  a cross-channel pattern that single-lead ECG classifiers cannot access.

\noindent\textbf{Leakage prevention.}
Each of the 498 records in the PhysioNet 2015 dataset represents a unique alarm event. Record identity was verified programmatically  all 498 record identifiers are distinct, confirming that no patient alarm event appears in both training and validation folds. Fold assignment is performed once prior to any model training using stratified sampling, and hyperparameter selection was conducted on a separate held-out split independent of the cross-validation procedure.

\subsection{Per-Alarm-Type Analysis}

\begin{figure}[!htb]
\centering
\includegraphics[width=\columnwidth]{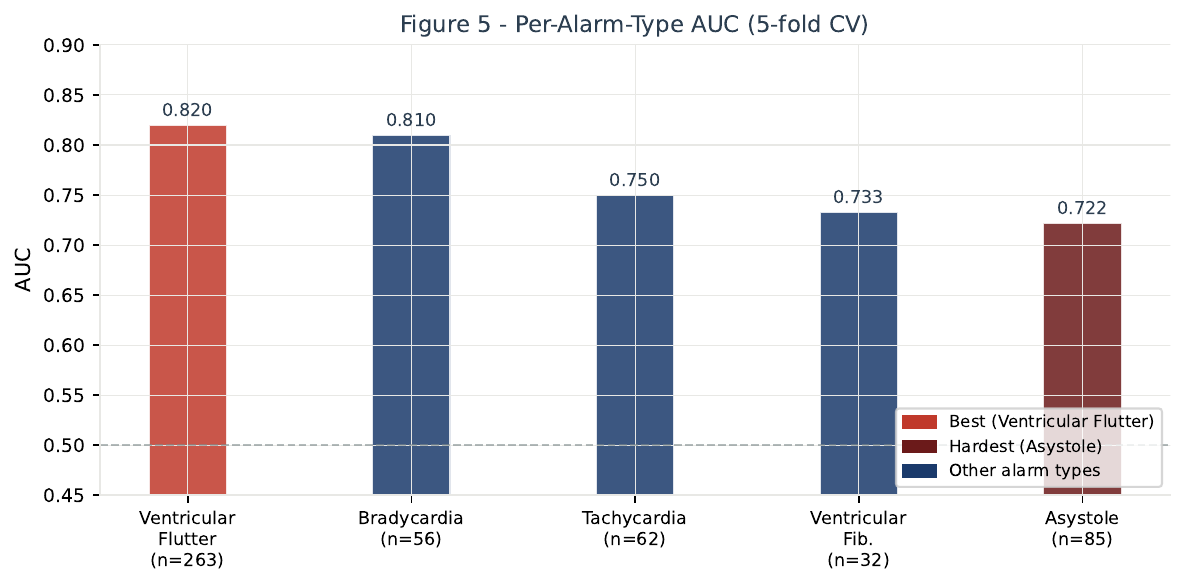}
\caption{Per-alarm-type AUC across five cross-validation
folds. Sample counts $n$ are shown below each alarm type.
Red bar indicates the best-performing type (Ventricular
Flutter); dark bar indicates the hardest (Asystole).}
\label{fig:peralarm}
\end{figure}

Table~\ref{tab:peralarm} and Figure~\ref{fig:peralarm}
report classification performance broken down by alarm
type across all five folds.

\begin{table}[!htb]
\centering
\caption{Per-alarm-type performance (5-fold CV).}
\label{tab:peralarm}
\begin{tabular}{lrrr}
\toprule
Alarm Type & $n$ & AUC & Accuracy \\
\midrule
Ventricular Flutter & 263 & 0.820 & 79.5\% \\
Bradycardia         &  56 & 0.810 & 69.6\% \\
Tachycardia         &  62 & 0.750 & 71.0\% \\
Ventricular Fib.    &  32 & 0.733 & 75.0\% \\
Asystole            &  85 & 0.722 & 76.5\% \\
\midrule
\textbf{Overall}
  & \textbf{498}
  & \textbf{0.815}
  & \textbf{76.5\%} \\
\bottomrule
\end{tabular}
\end{table}

\subsection{Error Analysis}

\begin{table}[!htb]
\centering
\caption{Clinical classification metrics, Experiment 04
(5-fold CV, threshold = 0.5).}
\label{tab:clinical}
\begin{tabular}{lr}
\toprule
Metric & Value \\
\midrule
Sensitivity (Recall) & 0.589 \\
Specificity          & 0.847 \\
Precision            & 0.641 \\
F1 Score             & 0.614 \\
NPV                  & 0.816 \\
Accuracy             & 0.765 \\
AUC                  & 0.822 \\
\midrule
True Positives  & 93  \\
True Negatives  & 288 \\
False Positives & 52  \\
False Negatives & 65  \\
\bottomrule
\multicolumn{2}{l}{\small True alarms: 158.
  False alarms: 340.}
\end{tabular}
\end{table}

Table~\ref{tab:clinical} reports clinical classification metrics at a decision threshold of 0.5. Sensitivity of 0.589 indicates the model correctly identifies 59\% of true alarms  a meaningful improvement over random
classification but insufficient for standalone clinical deployment. Specificity of 0.847 indicates the model correctly screens out 85\% of false alarms. The asymmetry between sensitivity and specificity reflects the class imbalance and the model's tendency to predict the majority
class (false alarm) under uncertainty. Threshold
optimization for clinical deployment would require prioritizing sensitivity at the cost of specificity, depending on the clinical risk tolerance of the deployment context.

Of 498 records, the model misclassifies 117 (23.5\%).
Errors break down as follows:

\begin{itemize}
  \item \textbf{False negatives} (65, 55.6\% of errors):
        true alarms classified as safe. These represent
        the most clinically dangerous error type, as a
        missed true alarm may result in delayed treatment.

  \item \textbf{False positives} (52, 44.4\% of errors):
        false alarms classified as real. These result in
        unnecessary clinical responses but do not directly
        endanger the patient.

  \item \textbf{High-confidence errors} (85, 72.6\% of
        errors): cases where the model assigned greater         than 80\% confidence to the incorrect class.
        Two representative cases: record a142s (Asystole,  99.35\% confidence, false negative) and record v143l (Ventricular Flutter, 99.48\% confidence, false negative). These suggest that certain true alarm signals share time-frequency characteristics with typical false alarm patterns  a limitation that patient-specific calibration may help address.
\end{itemize}

\subsubsection*{Class imbalance effects}

The 1:2 true-to-false alarm ratio (158 vs.\ 340) creates systematic pressure toward false alarm prediction. Despite class-weighted loss, false negatives outnumber false positives (65 vs.\ 52), suggesting the model still under-corrects for the minority class. Future work should explore oversampling techniques such as SMOTE or
alarm-type-specific decision thresholds.

\subsubsection*{Alarm-type difficulty}

Asystole alarms are the hardest to classify (AUC 0.722) despite being the most clinically urgent. The flatline pattern characteristic of true asystole can be reproduced by sensor disconnection  the most common source of false alarms  making signal-level separation difficult without additional contextual information. Ventricular Flutter achieves the highest AUC (0.820) and has the most
training examples (263), suggesting that data volume partially explains performance differences across alarm types.

\begin{figure}[!htb]
\centering
\includegraphics[width=\columnwidth]{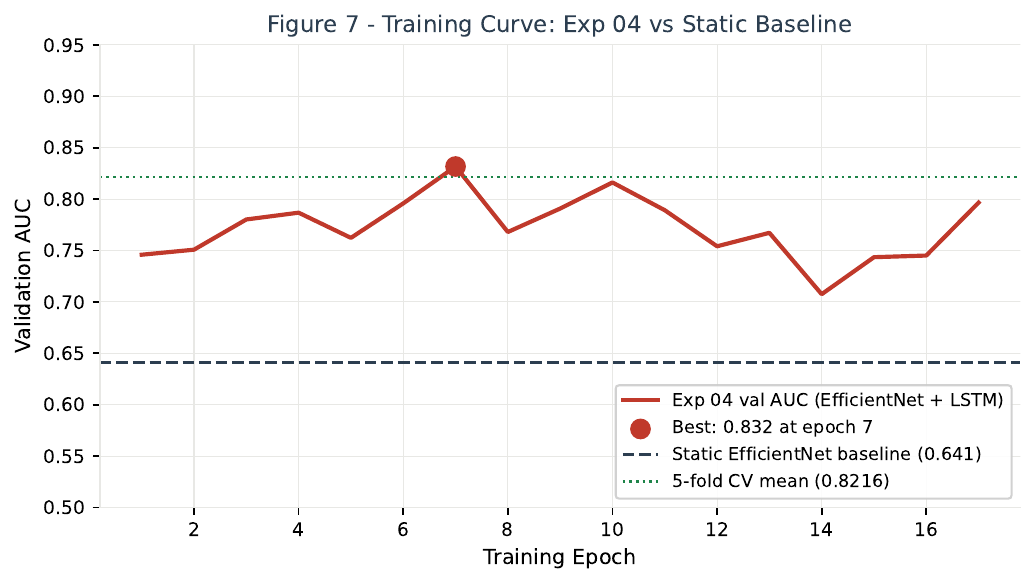}
\caption{Validation AUC per training epoch for Experiment
04 (EfficientNet + LSTM). The dashed line marks the static
EfficientNet baseline (0.641). The dotted line marks the
5-fold CV mean (0.8216). Early stopping prevents overfitting
beyond the best epoch.}
\label{fig:training}
\end{figure}

\section{Discussion}

The primary finding of this work is that temporal modeling of CWT scalogram sequences outperforms static classification on the ICU false alarm reduction task. The 18.1 AUC point
improvement from Experiment 01 to Experiment 04 reflects a meaningful difference in the information available to the model: a static classifier processes a single aggregated
representation of the 60-second window, while the LSTM processes how that representation evolves across six consecutive sub-windows.

The results are consistent with the hypothesis that the LSTM captures temporal degradation patterns that distinguish genuine arrhythmias from sensor artifacts, though it can be noted that this interpretation is observational. The internal representations of the LSTM have not been directly verified, and alternative explanations  such as the model benefiting from implicit data augmentation via chunk-level processing cannot be ruled out without further analysis.

The per-alarm-type results suggest that temporal modeling does not benefit all alarm types equally. Asystole remains the hardest alarm type despite the temporal architecture, likely because true asystole and sensor disconnection both produce flatline signals that are difficult to distinguish from 60 seconds of signal alone, regardless of temporal resolution. This suggests that the next meaningful improvement for Asystole classification may require patient-specific context rather than architectural changes.

\noindent\textbf{Comparison with prior work.}
Direct numerical comparison with published PhysioNet 2015 challenge results is not straightforward for two reasons.
First, prior challenge entries used full five-minute
retrospective recordings; our system uses only the final 60 seconds, a more constrained setting that reduces available context by 80\%. Second, challenge entries were scored using a custom sensitivity-specificity metric that penalizes missed true alarms more heavily than AUC does.
Within these constraints, our result of AUC 0.822 is
notable given the reduced signal window, and is competitive with the AUC ranges reported in post-challenge deep learning work~\cite{mousavi2020attention,zhou2022contrastive}.

\noindent\textbf{Clinical generalizability.}
This system was trained and evaluated on a single public dataset collected from a specific set of ICU monitors and patient populations. Generalization to different monitoring hardware, patient demographics, or clinical workflows cannot be assumed. External validation on independent
datasets is a necessary condition for any clinical
deployment consideration and is outside the scope of this work.

\section{Limitations}

\noindent\textbf{Dataset size.}
498 training records is small relative to the complexity of the EfficientNet-B0 + LSTM architecture. Results show variance across cross-validation folds (std 0.016), and individual training runs can differ by up to 4 AUC points due to random initialization. The cross-validation mean
is more reliable than any single run, but larger datasets would stabilize results and likely improve performance.

\noindent\textbf{Real-time constraint.}
Using only the final 60 seconds is clinically motivated but limits the temporal context available to the model. Prior work using five-minute windows has demonstrated that longer
context can improve classification, and the improvement from our temporal chunking may be partially attributable to making better use of a constrained window rather than temporal modeling per se.

\noindent\textbf{No patient-specific calibration.}
The model learns a population-level decision boundary. Individual patients have different baseline signal characteristics, and a model calibrated to a population mean may misclassify edge cases that would be clearly normal or abnormal for a specific patient. Patient-specific
calibration is a promising direction for improving both sensitivity and specificity.

\noindent\textbf{Class imbalance.}
The 1:2 true-to-false alarm ratio persists as a challenge despite class-weighted loss. The dominance of false negatives in the error analysis suggests that further work on minority class handling  including oversampling, threshold calibration, or alarm-type-specific models 
is warranted.

\noindent\textbf{Regulatory status.}
This system is a research prototype evaluated on a public benchmark dataset. Clinical deployment would require prospective validation on a larger independent dataset, regulatory clearance under FDA AI/ML Software as a Medical Device (SaMD) guidance, and integration with hospital monitoring infrastructure. No clinical claims are made.

\section{Conclusion}

This work represents SigmaMedStat, a temporal signal modeling framework for ICU false alarm reduction. By splitting 60-second physiological recordings into six consecutive 10-second
chunks, converting each chunk to a CWT scalogram, and processing the resulting sequence with a shared EfficientNet-B0 encoder and two-layer LSTM, we achieve mean AUC $0.822 \pm 0.016$ on five-fold stratified cross-validation  an improvement of 18.1 AUC points over a static EfficientNet baseline trained on the same
data and signal window.

Ablation studies confirm that both temporal chunking and multi-channel signal fusion contribute independently to this performance. Error analysis identifies false negatives
and high-confidence misclassifications as the primary failure modes, with Asystole remaining the hardest alarm type to classify.

The central finding is that the temporal trajectory of a physiological signal contains information that is not captured by static aggregation but is accessible to sequence models. Whether this reflects genuine learning of arrhythmia onset dynamics or a more general benefit
of sub-window processing requires further investigation. The observation may generalize to other clinical monitoring tasks where the pattern of signal change matters as much as its instantaneous state.

\section*{Acknowledgments}

The PhysioNet/Computing in Cardiology Challenge 2015 dataset was made available by Clifford et al.\ under open access terms. Experiments were conducted using PyTorch, torchvision, scikit-learn, and the PhysioNet \texttt{wfdb} library.


\end{document}